\RequirePackage[T1]{fontenc}
\documentclass[letterpaper,10pt,conference]{ieeeconf}
\IEEEoverridecommandlockouts
\usepackage{amsmath,amssymb}
\usepackage{graphicx}
\usepackage{booktabs,multirow}
\usepackage[table]{xcolor}
\usepackage{cite}
\usepackage[hidelinks,bookmarks=false,pdfversion=1.6]{hyperref}
\hypersetup{pdfauthor={Haozhe Lei and Sundeep Rangan},pdftitle={MAPLE-RF: Efficient Probabilistic RF Source Localization in Partially Explored Environments}}

\makeatletter
\renewcommand{\section}{\@startsection{section}{1}{\z@}%
  {1.2ex plus 0.5ex minus 0.2ex}{0.5ex plus 0.2ex}%
  {\normalfont\normalsize\centering\scshape}}
\renewcommand{\subsection}{\@startsection{subsection}{2}{\z@}%
  {1.2ex plus 0.5ex minus 0.2ex}{0.5ex plus 0.2ex}%
  {\normalfont\normalsize\itshape}}
\makeatother

\title{\LARGE\bfseries MAPLE-RF: Efficient Probabilistic RF Source Localization\\in Partially Explored Environments}
\author{Haozhe Lei$^{*}$ and Sundeep Rangan%
\thanks{All authors are with NYU WIRELESS, Tandon School of Engineering, New York University, Brooklyn, NY 11201, USA. Emails: \{hl4155,srangan\}@nyu.edu.}%
\thanks{$^{*}$Corresponding author: Haozhe Lei (hl4155@nyu.edu).}}

\begin{document}
\begin{NoHyper}
\maketitle

\begin{abstract}
Localizing a radio-frequency (RF) transmitter from received signals often requires a model of the environment to predict how obstacles block and reflect the signal. In many robotic applications, however, only a partial map is available, particularly when a robot localizes the source while exploring with simultaneous localization and mapping (SLAM). We study single-snapshot transmitter localization on such partially explored maps and compare two approaches that output a posterior over transmitter locations. The first extends a digital-twin method, which ray-traces every candidate location, to partial maps by treating unexplored space as free and training on mixed map coverage. The second, MAPLE-RF, encodes estimated path angles of arrival and signal-to-noise ratios as grid channels aligned with map knownness, occupancy, and line-of-sight visibility, and a U-Net scores all candidate positions in one pass without simulating propagation at inference. Ray-tracing simulations of indoor environments indicate that training on mixed map coverage is essential for both approaches. The digital-twin approach is more accurate on most single-snapshot metrics, while MAPLE-RF comes close at a query cost that does not depend on the propagation model and is more than two orders of magnitude below a fresh full-grid query with general-purpose ray tracing. Both outperform Gaussian and Gaussian-mixture baselines, and on exploration routes guided by its own estimates, fused MAPLE-RF posteriors place more probability near the source than the compared methods. Code and data will be released.
\end{abstract}


\section{Introduction}
\label{sec:introduction}

\makeatletter\global\@topnum=-1\relax\makeatother
\begin{figure}[!t]
    \centering
    \includegraphics[width=0.97\columnwidth]{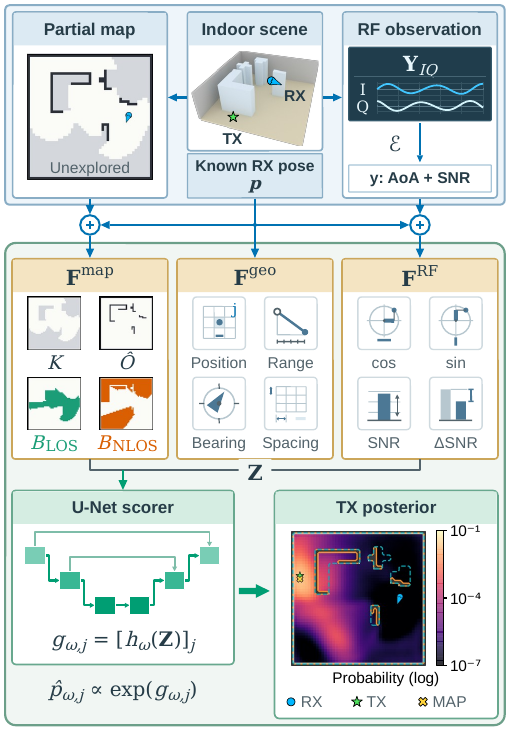}
    \makeatletter
    \long\def\@makecaption#1#2{%
        \@IEEEfigurecaptionsepspace
        \parbox[t]{\hsize}{\footnotesize\noindent\mbox{#1.\hspace{0.5em}}#2}}
    \makeatother
    \caption{Illustration of MAPLE-RF for single-snapshot RF source localization with a partial map. One receiver (RX) provides its known pose and multipath angle-of-arrival (AoA) and signal-to-noise ratio (SNR) estimates. Map/visibility, geometry, and RF features are concatenated as $\mathbf Z$. A residual U-Net scores all candidate transmitter (TX) positions in one pass; normalization yields the position posterior. The non-line-of-sight (NLOS) example uses Spatial supervision and Robust training. Gray marks unexplored space; green/orange indicate confirmed line-of-sight (LOS)/NLOS. Solid orange and dashed cyan contours outline observed and true obstacles. In-phase and quadrature (IQ) traces and feature icons are schematic. MAP denotes maximum a posteriori.}
    \label{fig:method_overview}
\end{figure}

Localizing a radio-frequency (RF) transmitter (TX) from signals observed at a receiver (RX) is a basic task in indoor positioning~\cite{SpotFi2015,DLoc2020}, wireless-guided robot navigation~\cite{Yin2022,Yin2024}, and robotic search for radio sources~\cite{Twigg2012,Denniston2023}. Obstacles often block the direct path and create reflections, and environment geometry helps interpret these paths. Indeed, learned models predict radio maps from scene geometry~\cite{RadioUNet2021}, and such maps support localization~\cite{LocUNet2023}. Ray-tracing digital twins (DTs)~\cite{SionnaRT2025} simulate wireless environments for training navigation policies~\cite{Li2025DTWIN} and can predict the multipath that each candidate TX location would produce~\cite{LOCUSDT}. These methods typically assume an accurate map of the full environment.

A robot entering an unfamiliar building rarely has such a map. It builds one while exploring~\cite{Yamauchi1997}, using simultaneous localization and mapping (SLAM)~\cite{Cadena2016}, and large regions remain unobserved during the search. These regions may hide obstacles that block or reflect the received paths, so a localizer must distinguish known free space from unexplored space. Two routes are available. A DT can ray-trace the observed map, treating unexplored space as free, but must repeat its per-candidate simulation whenever the map or RX pose changes. Alternatively, a network can learn to interpret RF paths in the context of the partial map, without simulating propagation at inference. This raises two questions: how should RF observations be combined with a partial, evolving map, and when is per-candidate propagation simulation worth its cost?

With one RX, one snapshot, and a partial map, many TX locations can be consistent with the data. An angle of arrival (AoA) constrains the bearing more tightly than the range, a reflected path can be explained by different surfaces, and an unexplored region can hide the true propagation path. A point estimate conceals this ambiguity. Following probabilistic RF localization~\cite{Gonultas2022,Leitinger2019,MCCLE2025}, we therefore estimate a posterior over candidate TX positions. The posterior indicates where the source may be and how concentrated that belief is, and posteriors from successive observations can be fused as the robot moves.

\noindent\textbf{Contributions.}
We study single-snapshot RF source localization on partially explored maps and compare our LOCUS-DT extension with our proposed MAPLE-RF (Fig.~\ref{fig:method_overview}).

\begin{itemize}
\item \emph{Problem and benchmark.} We formulate the task as estimating a posterior over candidate TX positions from estimated multipath, the partial map, and a known RX pose. We will release a benchmark of 2,400 indoor layouts ray-traced at 10\,GHz with direct and first-order specular paths, containing 115,200 RF observations with multipath descriptors estimated from simulated in-phase and quadrature (IQ) samples, complete maps, and 1,512 partial maps from simulated lidar exploration at three coverage levels.
\item \emph{Partial-map extension of DT localization.} We extend LOCUS-DT~\cite{LOCUSDT}, which compares observed paths with DT-predicted paths for each candidate in a known environment, to partial maps (Section~\ref{sec:dt_extension}): the DT ray-traces the observed occupancy with unexplored space treated as free, the scorer receives line-of-sight/non-line-of-sight (LOS/NLOS) visibility features, and training uses mixed map coverage. Relative to training on complete maps only, mixed-coverage training raises partial-map 1\,m recall from 54.7\% to 66.4\%.
\item \emph{MAPLE-RF: learned localization without propagation simulation at inference.} Estimated path AoAs and signal-to-noise ratios (SNRs) become per-cell channels encoding each path's angular compatibility with the candidate bearing, stacked with knownness, occupancy, and confirmed LOS/NLOS channels. A standard residual U-Net scores all candidates jointly, producing a nonparametric posterior that can follow obstacle geometry and contain several modes. With mixed-coverage training, MAPLE-RF-S retains about 94\% of its complete-map 1\,m recall when 75--85\% of the map is unobserved; on partial maps, visibility channels improve all six metrics under this training but degrade quality under complete-map-only training. With mixed coverage, a distance-based Spatial target also improves all six quality metrics over one-hot supervision.
\item \emph{Comparison of the two approaches.} With mixed-coverage training, full-grid DT leads on four of six metrics averaged over partial maps, including 1\,m recall (66.4\% vs.\ 63.9\%) and mass negative log-likelihood (Mass NLL), while MAPLE-RF-S has 7.8\% lower expected distance and nearly identical maximum a posteriori (MAP) error. MAPLE-RF's query cost does not depend on the propagation model; given estimated RF paths, MAPLE-RF-S is about $218\times$ faster per fresh query than full-grid DT localization with general-purpose ray tracing, although a solver specialized to our first-order paths would narrow this gap. It remains about $25\times$ faster than an inference-only $13\times13$ DT grid while improving all six metrics. When posteriors are fused along 100 initially NLOS exploration sequences, with routes selected by the MAPLE-RF-S MAP estimate (a protocol that may favor MAPLE-RF-S), accumulated MAPLE-RF-S posteriors have the highest mean probability within 1\,m of the source at every step. Both approaches outperform Gaussian and Gaussian-mixture heads that share MAPLE-RF's encoder architecture in localization quality.
\end{itemize}

\noindent\textbf{Scope.} The evaluation is simulation-only, with one room size, polygonal obstacles, first-order reflections, and noise-free observed maps and RX poses (Section~\ref{sec:conclusion}).

\par\noindent\textbf{\mbox{Related Work.}}
Geometric RF localization combines angle and delay estimates with direct-path selection across multiple access points~\cite{SpotFi2015}, while supervised models learn localization from RF measurements collected and labeled by a mapping robot~\cite{DLoc2020}. Robotic search for radio sources has used sequential signal-strength or range measurements, in known~\cite{Charrow2014} and unknown environments~\cite{Twigg2012,Denniston2023,Kim2025}; such searchers could use single-snapshot posteriors as measurement models. Propagation models and physical priors also guide wireless robot navigation~\cite{Yin2024,Li2025DTWIN,Li2025PiPRL}, and signal strength has been predicted from geometry that robots build online~\cite{Clark2022}. Multipath-assisted localization typically assumes a known floor plan or DT~\cite{Kanhere2025,LOCUSDT}, or estimates the environment jointly from radio measurements along a trajectory~\cite{Leitinger2019}, which can also accommodate inaccurate floor plans~\cite{Leitinger2015}. Probabilistic localization retains competing hypotheses: position distributions from channel state information can be fused across antennas and access points~\cite{Gonultas2022}, and Gaussian and mixture-density position models have been used in multipath-assisted fingerprinting~\cite{Ulmschneider2023,Ulmschneider2024}. For single snapshots, MC-CLE scores candidate positions from AoAs and SNRs under LOS propagation~\cite{MCCLE2025}, and LOCUS-DT extends this to multipath with DT-predicted paths in a known environment~\cite{LOCUSDT}. To our knowledge, localizing a static TX from a single multipath snapshot conditioned on a partially explored occupancy map, one that distinguishes known-free, known-occupied, and unexplored space, has received little attention.

\section{Problem Formulation}
\label{sec:problem}

\subsection{Localization Setting}
\label{sec:localization_setting}

Consider a TX at an unknown, fixed position $\mathbf x_0^t\in\mathbb R^2$ in a global coordinate system. We use $\mathbf x^t$ for a generic TX position; superscripts $t$ and $r$ denote TX and RX quantities, respectively. A query takes map and visibility information $\mathcal M$, RF observation $\mathbf y$, and known RX pose $\mathbf p=(\mathbf x^r,\phi^r)$, with position $\mathbf x^r\in\mathbb R^2$ and heading $\phi^r\in[-\pi,\pi)$. SLAM~\cite{Cadena2016} can provide a partial map and the RX pose in a common coordinate frame.

The TX antenna is assumed isotropic, and the RX uses a directional array oriented by $\phi^r$. Unobserved geometry, the RX antenna's directional response, and uncertainty in RF observations can leave several TX locations consistent with one snapshot. We therefore seek the TX-position posterior \mbox{$p(\mathbf x^t\mid\mathcal M,\mathbf p,\mathbf y)$}.

\subsection{RF Observation Interface}
\label{sec:observation}

To form $\mathbf y$, the RX collects IQ samples $\mathbf Y_{\mathrm{IQ}}$ from a transmitted waveform. An estimator $\mathcal E$ extracts a ranked multipath description, with $\mathbf r_k$ denoting the descriptor of retained path $k$:
\begin{equation}
\mathbf y=\mathcal E(\mathbf Y_{\mathrm{IQ}})
         =(\mathbf r_1,\ldots,\mathbf r_{\kappa_y}),
\quad \mathbf r_k=(\widehat\theta_k,\widehat\gamma_k).
\label{eq:observation}
\end{equation}
Here, $\kappa_y$ is the number of retained descriptors, and $\widehat\theta_k$ and $\widehat\gamma_k$ are the estimated AoA and path SNR, respectively. The descriptors are ordered by decreasing estimated SNR. Each AoA describes the apparent arrival direction of a path, expressed in the world frame using the RX heading.

\subsection{Map and Visibility Information}
\label{sec:map_information}

Let $O$ denote the true occupancy map and $K$ the observed-region mask. The observed occupancy is $\widehat O=K\odot O$, where $\odot$ denotes pointwise multiplication. The pair $(K,\widehat O)$ distinguishes occupied, free, and unexplored space~\cite{Yamauchi1997}.

To complement the RF descriptors, we encode LOS and NLOS link states, also used in wireless navigation~\cite{Yin2022,Yin2024,Li2025DTWIN}. The binary fields $B_{\mathrm{LOS}}$ and $B_{\mathrm{NLOS}}$, derived from $(K,\widehat O,\mathbf p)$, mark candidate locations with confirmed clear and blocked direct paths to the RX, respectively. Both fields are zero where visibility is unknown. Together, these four fields form $\mathcal M=(K,\widehat O,B_{\mathrm{LOS}},B_{\mathrm{NLOS}})$.

\section{Methodology}
\label{sec:methodology}

To estimate the TX-position posterior, we develop \emph{Map-Aware Probabilistic Location Estimation from Radio-Frequency Observations} (MAPLE-RF) and extend LOCUS-DT to partial maps. We first describe MAPLE-RF, which combines partial-map information, RF descriptors, and RX-relative geometry on a common grid to score all candidates jointly.

\subsection{Grid Posterior and Supervision}
\label{sec:score_model}
\label{sec:loss_design}

We represent the TX-position posterior by cell masses on a fixed uniform $H\times W$ grid, with interior support $\mathcal C$ and cell representatives $\bar{\mathbf x}_j^t$. Given $(\mathcal M,\mathbf p,\mathbf y)$, the network with trainable parameters $\boldsymbol\omega$ jointly predicts scores $\mathbf g_{\boldsymbol\omega}=(g_{\boldsymbol\omega,j})_{j\in\mathcal C}$. The cell-mass vector is $\widehat{\mathbf p}_{\boldsymbol\omega}=\operatorname{softmax}_{\mathcal C}(\mathbf g_{\boldsymbol\omega})$. The support remains fixed across map views; occupancy and visibility condition the scores. Summing cell masses over a region approximates its location probability. The MAP estimate is the representative of the highest-mass cell: $\widehat{\mathbf x}^t=\bar{\mathbf x}_{\hat j}^t$, where $\hat j=\arg\max_{j\in\mathcal C}\widehat p_{\boldsymbol\omega,j}$.

Each training example contains $(\mathcal M,\mathbf p,\mathbf y,\mathbf x_0^t)$. Our MAPLE-RF-H and MAPLE-RF-S variants use the same inputs and network architecture, differing only in their Hard/Spatial targets $q$. Both minimize the cross-entropy loss:
\begin{flalign}
&\mathcal L(\boldsymbol\omega)
\!=\!\!\frac{1}{N}\sum_{n=1}^{N}\!\left[
-\!\sum_{j\in\mathcal C}\!q_{n,j}g_{\boldsymbol\omega,n,j}
\!+\!\log\!\sum_{i\in\mathcal C}\!\exp(g_{\boldsymbol\omega,n,i})
\right], &&
\label{eq:training_loss}
\raisetag{24pt}
\end{flalign}
where $N$ is the number of training examples and $g_{\boldsymbol\omega,n,j}$ is the score at cell $j$ for example $n$.

\textbf{Hard target (H).} All target mass is assigned to the grid point nearest to the true TX position: $q_j^{\mathrm H}=\mathbf 1\{j=j_0\}$, with
$j_0=\arg\min_{j\in\mathcal C}\|\bar{\mathbf x}_j^t-\mathbf x_0^t\|_2$.

\textbf{Spatial target (S).} Target mass is distributed according to physical distance from the continuous TX position:
\begin{equation}
q_j^{\mathrm S}
=\frac{\exp\!\left(-\|\bar{\mathbf x}_j^t-\mathbf x_0^t\|_2^2/(2\sigma_{\mathrm{sp}}^2)\right)}
{\sum_{i\in\mathcal C}\exp\!\left(-\|\bar{\mathbf x}_i^t-\mathbf x_0^t\|_2^2/(2\sigma_{\mathrm{sp}}^2)\right)}.
\label{eq:spatial_target}
\end{equation}
Here, $\sigma_{\mathrm{sp}}$ sets the target's spatial spread in meters.
The target varies smoothly as the TX moves within a cell and gives neighboring cells similar supervision.

\subsection{Input Representation}
\label{sec:input_representation}

Stacking the four fields of $\mathcal M$ gives the binary tensor $\mathbf F^{\mathrm{map}}\in\{0,1\}^{4\times H\times W}$.
The pair $(K(j),\widehat O(j))$ distinguishes known free $(1,0)$, known occupied $(1,1)$, and unexplored $(0,0)$ cells. For an interior cell not known to be occupied, $B_{\mathrm{NLOS}}(j)=1$ when an observed obstacle blocks the RX-to-cell segment, whereas $B_{\mathrm{LOS}}(j)=1$ when the entire segment is observed and free. Both channels are zero otherwise. Fig.~\ref{fig:map_visibility} illustrates the four channels at different map coverage levels.

\begin{figure}[!t]
    \centering
    \includegraphics[width=0.97\columnwidth]{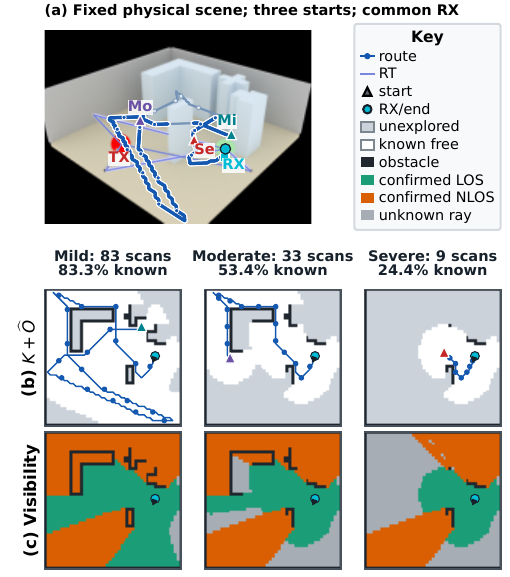}
    \caption{Map and visibility information for a common RX position. (a) Three exploration routes in the same physical scene end at the RX; propagation paths are obtained by ray tracing (RT). Mi/Mo/Se denote Mild/Moderate/Severe route starts. (b) Knownness and occupancy grids. (c) Confirmed LOS/NLOS and unknown visibility. The coverage levels follow the definition in Section~\ref{sec:map_dataset}.}
    \label{fig:map_visibility}
\end{figure}

For the geometry tensor $\mathbf F^{\mathrm{geo}}\in\mathbb R^{9\times H\times W}$, let $\boldsymbol\delta_j=\bar{\mathbf x}_j^t-\mathbf x^r$ and $\beta_j=\operatorname{atan2}(\delta_{j,y},\delta_{j,x})$. With room side lengths $L_x,L_y$ and horizontal lower bound $x_{\min}$, these channels contain: candidate coordinates scaled to $[-1,1]$; offsets $\delta_{j,x}/L_x$ and $\delta_{j,y}/L_y$; normalized distance $\sqrt{(\delta_{j,x}/L_x)^2+(\delta_{j,y}/L_y)^2}/\sqrt{2}$; $\cos(\beta_j-\phi^r)$ and $\sin(\beta_j-\phi^r)$; and $\log(\Delta_x/1\,\mathrm m)$ and $\log(\Delta_y/1\,\mathrm m)$ for grid spacings $\Delta_x,\Delta_y$. The normalized horizontal coordinate, for example, is $2(\bar x_{j,x}^t-x_{\min})/L_x-1$.

The RF tensor $\mathbf F^{\mathrm{RF}}\in\mathbb R^{4\kappa\times H\times W}$ uses $\kappa$ fixed descriptor slots, with padding when $\kappa_y<\kappa$. Its four-channel block for slot $k$ at cell $j$ is
\begin{equation}
\begin{aligned}
&\mathbf F_{j,k}^{\mathrm{RF}}=\big[
\cos(\beta_j-\widehat\theta_k),\ \sin(\beta_j-\widehat\theta_k),\\
&\quad\tanh(\widehat\gamma_k/(20\,\mathrm{dB})),\
\tanh((\widehat\gamma_k-\widehat\gamma_1)/(20\,\mathrm{dB}))\big]^\top.
\end{aligned}
\label{eq:path_features}
\end{equation}
SNR is expressed in decibels. The two pairs encode angular compatibility with the candidate bearing and absolute/relative path strength, respectively.

\subsection{Neural Realization}
\label{sec:network}

The scorer $h_{\boldsymbol\omega}$ receives the channel-wise concatenation $\mathbf Z=[\mathbf F^{\mathrm{map}};\mathbf F^{\mathrm{geo}};\mathbf F^{\mathrm{RF}}]\in\mathbb R^{(13+4\kappa)\times H\times W}$.

The scorer in Fig.~\ref{fig:method_overview} is a residual U-Net~\cite{UNet2015} with widths 64, 128, and 256 and two residual blocks per encoder level. The blocks use group normalization and sigmoid linear unit (SiLU) activations. Strided convolutions downsample; upsampling and convolution restore resolution with concatenated encoder--decoder skips. Two bottleneck blocks include spatial self-attention~\cite{Attention2017}. A final convolution produces $h_{\boldsymbol\omega}(\mathbf Z)\in\mathbb R^{H\times W}$ in one pass, with $g_{\boldsymbol\omega,j}=[h_{\boldsymbol\omega}(\mathbf Z)]_j$.

\subsection{Partial-Map Extension of LOCUS-DT}
\label{sec:dt_extension}

We extend LOCUS-DT~\cite{LOCUSDT}, which assumes a known environment, to partial maps. Its neural scorer compares ideal, noise-free DT multipath predictions for each candidate with the observed paths. We run the DT on the observed occupancy $\widehat O$ at RX pose $\mathbf p$, so unexplored cells are treated as free, and normalize scores over interior candidates with $\widehat O(j)=0$. For the three SNR-ranked slots in each profile, sine/cosine AoA encodings, scaled SNRs, and SNR differences between corresponding ranks give $21$ features, which we augment with $B_{\mathrm{LOS}}(j)$ and $B_{\mathrm{NLOS}}(j)$. Mixed-coverage training uses DT predictions from incomplete geometry (Section~\ref{sec:training_protocol}). Unlike MAPLE-RF, this approach must ray-trace every candidate whenever the map or RX pose changes.

\section{Experimental Setup}
\label{sec:experimental_setup}

\begin{figure*}[!t]
    \centering
    \includegraphics[width=0.99\textwidth]{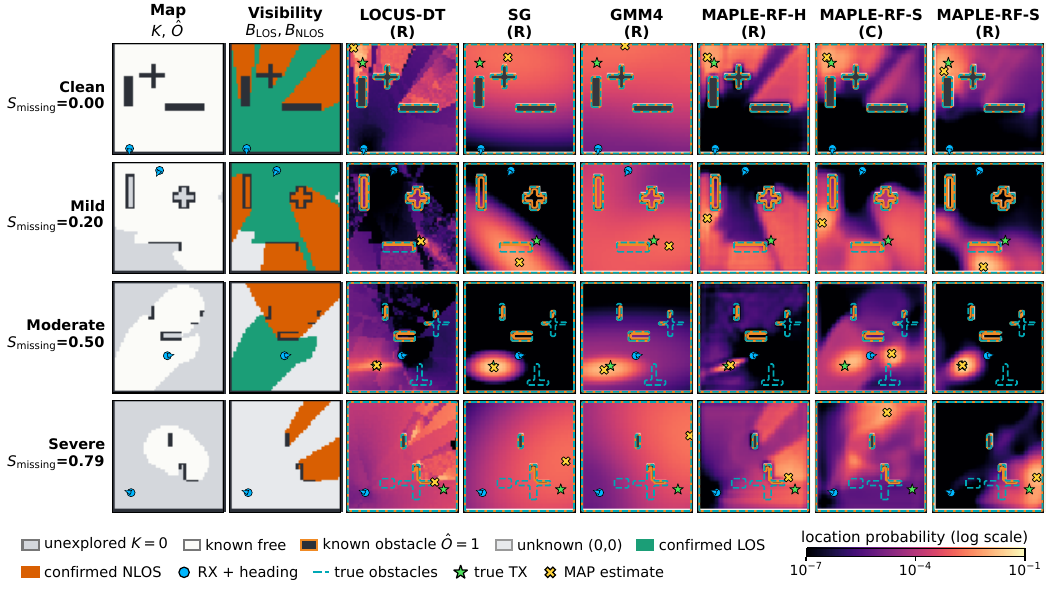}
    \caption{Single-snapshot localization examples across map coverage. The first two columns show the available map and derived visibility; the remaining columns show estimated location probabilities on a common logarithmic color scale. Stars and crosses indicate the true TX and MAP estimates. Solid orange and dashed cyan contours outline observed and true obstacles. H/S denote Hard/Spatial supervision, and C/R denote Clean/Robust training.}
    \label{fig:localization_examples}
\end{figure*}

\begin{figure}[!t]
    \centering
    \includegraphics[width=0.97\columnwidth]{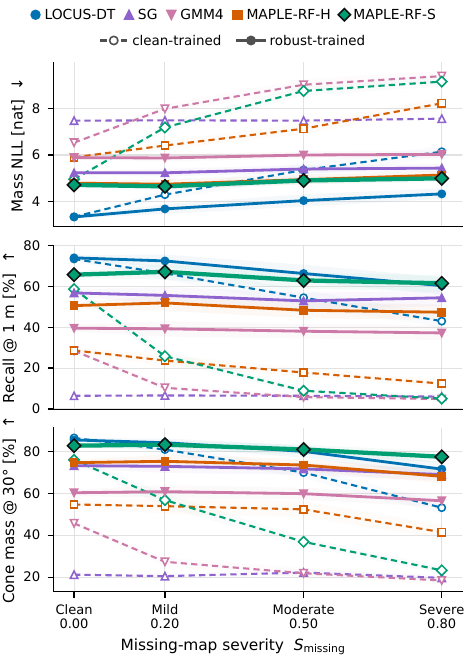}
    \caption{Single-snapshot performance across map coverage. Lines show means; shaded bands show 95\% confidence intervals from 2,000 layout-family bootstrap resamples. Each layout--map unit averages 48 RF snapshots. Clean uses 240 units; each incomplete-map level uses its own 48-layout subset. H/S denote Hard/Spatial supervision; solid/dashed lines denote Robust/Clean training.}
    \label{fig:map_coverage_results}
\end{figure}

\subsection{Simulation and Measurement Settings}
\label{sec:simulation_settings}

We generate $10\,\mathrm{m}\times10\,\mathrm{m}$ rooms with one to four reflective obstacles: horizontal/vertical bars, T and L shapes, and crosses. NVIDIA Sionna RT~\cite{SionnaRT2025}, a ray-tracing (RT) simulator, models $10\,\mathrm{GHz}$ direct paths and first-order specular reflections. The TX uses a single isotropic antenna and the RX an eight-element directional array, both at $0.615\,\mathrm{m}$ height. Bandwidth, transmit power, and receiver noise figure are $200\,\mathrm{MHz}$, $10\,\mathrm{dBm}$, and $7\,\mathrm{dB}$, respectively. A fixed $49\times49$ uniform grid spans the room at approximately $0.208\,\mathrm{m}$ spacing; excluding its boundary cells leaves $47\times47=2{,}209$ candidates for all MAPLE-RF training and inference.

Estimator $\mathcal E$ uses the iterative Levenberg--Marquardt method of LOCUS-DT~\cite{LOCUSDT,Bomfin2025}, processing $128$ frequency bins and estimating up to $16$ paths. We retain up to $\kappa=3$ AoA--SNR pairs in decreasing SNR order, with a $-65\,\mathrm{dB}$ SNR floor. Missing slots use seeded random AoAs and the floor.

\subsection{Map Coverage and Dataset Construction}
\label{sec:map_dataset}

Motivated by range-sensor SLAM~\cite{Cadena2016}, we simulate exploration with $360^\circ$ scans limited to $1.8\,\mathrm{m}$. Rays mark traversed free cells and the first occupied cell as known; cells behind remain unknown. Each snapshot yields $K$ and $\widehat O$ with a known outer room boundary (Fig.~\ref{fig:map_visibility}).

Using known/unknown cells~\cite{Yamauchi1997}, we measure map incompleteness by the unobserved fraction $S_{\mathrm{missing}}=1-|\mathcal C|^{-1}\sum_{j\in\mathcal C}K(j)$ on the fixed interior grid.
Mild, Moderate, and Severe maps have $S_{\mathrm{missing}}\in[0.15,0.25]$, $[0.45,0.55]$, and $[0.75,0.85]$, respectively; Clean maps have $S_{\mathrm{missing}}=0$. Connected known-region masks augment training. Clean (C) training uses complete maps; Robust (R) training uses mixed coverage.

Layouts are split before map augmentation (Table~\ref{tab:dataset_summary}). Each layout has two RX poses, each paired with $24$ TX positions, yielding $115{,}200$ RF observations. Map variants preserve the physical scene, TX--RX configurations, and RF observations; methods share RF--map pairings within each training regime.

\begin{table}[tb]
\centering
\caption{Dataset split and available RF--map examples}
\label{tab:dataset_summary}
\small
\setlength{\tabcolsep}{2.5pt}
\begin{tabular*}{\columnwidth}{@{\extracolsep{\fill}}lrrrr@{}}
\toprule
Quantity & Train & Val. & Test & Total \\
\midrule
Layouts & 1,920 & 240 & 240 & 2,400 \\
Partial maps & 1,272 & 96 & 144 & 1,512 \\
RF--map examples & 153,216 & 16,128 & 18,432 & 187,776 \\
\bottomrule
\end{tabular*}
\par\smallskip
\begin{minipage}{\columnwidth}\footnotesize
Each layout has one Clean map. Partial maps are balanced across the three severity levels. Each map is paired with its layout's 48 RF observations.
\end{minipage}
\end{table}

\subsection{Model Variants and Baselines}
\label{sec:comparison_methods}

\emph{LOCUS-DT} denotes the partial-map DT approach of Section~\ref{sec:dt_extension}, evaluated on the same observed maps and RF snapshots as MAPLE-RF. LOCUS-DT (C), trained only on complete maps as the original method assumes, shows the effect of ignoring map incompleteness during training. For reduced inference cost, we form $25\times25$ and $13\times13$ RT grids by retaining every second or fourth node along each axis. Both reuse the trained R scorer and the same test RF snapshots, maps, and propagation settings. We linearly interpolate unnormalized scores onto the original $49\times49$ grid, extend them by the nearest candidate outside the convex hull, and apply softmax over the original LOCUS-DT support. All quality metrics, including Mass NLL, use the original grid without resnapping the truth to a coarse grid.

We adapt Gaussian baselines from MC-CLE and LOCUS-DT~\cite{MCCLE2025,LOCUSDT}, also used in RF localization~\cite{Ulmschneider2023,Ulmschneider2024}. The \emph{single Gaussian (SG)} and \emph{four-component Gaussian mixture model (GMM4)} use full covariances. Both share MAPLE-RF's full input $\mathbf Z$ and encoder architecture, including bottleneck attention, followed by global average pooling and a distribution-parameter head. Both use the Spatial target~\eqref{eq:spatial_target} and evaluate and normalize densities over the same fixed interior support.

\subsection{Training Protocol}
\label{sec:training_protocol}

We train all five models under C and R; MAPLE-RF-S (R) is our main configuration. C uses all Clean training examples. R samples $55\%$ Clean maps, $40\%$ exploration snapshots, and $5\%$ connected-mask augmentation from the full training pool, balancing each partial-map category across severity levels. Both settings use $92{,}160$ RF--map examples per epoch, pairing one map view per training layout with all its RF observations.

MAPLE-RF, SG, and GMM4 use Adam with a peak learning rate of $0.0012$, $90$ warmup updates, and cosine decay. Two-step accumulation of $1{,}024$-example batches gives an effective batch size of $2{,}048$. Training lasts at most $100$ epochs ($4{,}500$ updates), with $\sigma_{\mathrm{sp}}=0.3125\,\mathrm{m}$ in \eqref{eq:spatial_target}. LOCUS-DT uses AdamW with learning rate $0.005$, weight decay $10^{-4}$, and batch size $2{,}048$, for at most $100$ epochs. All models use the same training seed. Checkpoint selection and early stopping use validation Mass NLL.

\providecommand{\meanstd}[2]{#1\,\pm\,#2}
\begin{table*}[t]
\centering
\footnotesize
\setlength{\tabcolsep}{1.5pt}
\renewcommand{\arraystretch}{0.96}
\definecolor{mapleproposed}{RGB}{237,246,231}
\definecolor{maplebest}{RGB}{0,78,132}
\definecolor{maplesecond}{RGB}{151,67,0}
\caption{Single-snapshot distribution quality, MAP localization summaries, and query cost on partially explored maps}
\label{tab:main_results_quality_cost}
\begin{tabular}{p{.125\textwidth}p{.045\textwidth}*{2}{>{\centering\arraybackslash}p{.11\textwidth}}>{\centering\arraybackslash}p{.115\textwidth}*{3}{>{\centering\arraybackslash}p{.11\textwidth}}>{\centering\arraybackslash}p{\dimexpr.165\textwidth-18\tabcolsep\relax}}
\toprule
 & & \multicolumn{4}{c}{Full distribution} & \multicolumn{2}{c}{MAP summaries} & Cost \\
\cmidrule(lr){3-6}\cmidrule(lr){7-8}\cmidrule(lr){9-9}
Model & Train & \shortstack{Mass NLL\\{[nat]} $\downarrow$} & \shortstack{ES\\{[m]} $\downarrow$} & \shortstack{Expected\\distance [m] $\downarrow$} & \shortstack{1\,m mass\\{[\%]} $\uparrow$} & \shortstack{Error\\{[m]} $\downarrow$} & \shortstack{1\,m recall\\{[\%]} $\uparrow$} & \shortstack{Query time\\{[ms]} $\downarrow$} \\
\midrule
LOCUS-DT & C & $5.26 \pm 1.19$ & $1.15 \pm 0.52$ & $2.21 \pm 0.77$ & $42.6 \pm 15.9$ & $1.75 \pm 0.81$ & $54.7 \pm 15.6$ & $3116.0 \pm 403.5$ \\
LOCUS-DT & R & \textcolor{maplebest}{$\mathbf{4.01 \pm 0.88}$} & \textcolor{maplebest}{$\mathbf{0.78 \pm 0.35}$} & \textcolor{maplesecond}{$1.68 \pm 0.63$} & \textcolor{maplebest}{$\mathbf{57.0 \pm 13.0}$} & \textcolor{maplesecond}{$1.32 \pm 0.73$} & \textcolor{maplebest}{$\mathbf{66.4 \pm 14.7}$} & $3116.0 \pm 403.5$ \\
LOCUS-DT $25^2$ & R & \textcolor{maplesecond}{$4.55 \pm 0.75$} & $0.85 \pm 0.34$ & $1.86 \pm 0.61$ & $52.2 \pm 12.3$ & $1.36 \pm 0.69$ & \textcolor{maplesecond}{$65.4 \pm 14.3$} & $902.1 \pm 111.7$ \\
LOCUS-DT $13^2$ & R & $5.45 \pm 0.57$ & $1.04 \pm 0.32$ & $2.26 \pm 0.56$ & $40.1 \pm 9.9$ & $1.60 \pm 0.66$ & $54.1 \pm 13.6$ & $353.9 \pm 75.3$ \\
\midrule
SG & C & $7.52 \pm 0.13$ & $2.30 \pm 0.18$ & $4.59 \pm 0.21$ & $4.0 \pm 0.5$ & $3.86 \pm 0.46$ & $6.4 \pm 3.5$ & \textcolor{maplesecond}{$11.6 \pm 5.6$} \\
SG & R & $5.35 \pm 0.61$ & $0.95 \pm 0.31$ & $1.90 \pm 0.50$ & $39.5 \pm 9.8$ & $1.48 \pm 0.58$ & $54.4 \pm 14.5$ & \textcolor{maplesecond}{$11.6 \pm 5.6$} \\
GMM4 & C & $8.82 \pm 0.96$ & $2.76 \pm 0.53$ & $4.11 \pm 0.49$ & $5.5 \pm 2.1$ & $3.90 \pm 0.64$ & $7.0 \pm 5.4$ & \textcolor{maplebest}{$\mathbf{10.8 \pm 4.9}$} \\
GMM4 & R & $5.97 \pm 0.54$ & $1.34 \pm 0.35$ & $2.66 \pm 0.54$ & $24.7 \pm 6.9$ & $2.25 \pm 0.75$ & $38.2 \pm 13.0$ & \textcolor{maplebest}{$\mathbf{10.8 \pm 4.9}$} \\
\midrule
MAPLE-RF-H & C & $7.26 \pm 1.06$ & $2.06 \pm 0.55$ & $3.33 \pm 0.54$ & $16.7 \pm 5.9$ & $3.44 \pm 0.74$ & $18.0 \pm 8.9$ & $14.2 \pm 4.4$ \\
MAPLE-RF-H & R & $4.94 \pm 0.74$ & $1.09 \pm 0.32$ & $2.10 \pm 0.55$ & $42.5 \pm 11.5$ & $1.94 \pm 0.68$ & $49.2 \pm 13.3$ & $14.2 \pm 4.4$ \\
MAPLE-RF-S & C & $8.38 \pm 1.14$ & $2.47 \pm 0.60$ & $3.73 \pm 0.77$ & $11.6 \pm 8.2$ & $3.84 \pm 0.98$ & $13.3 \pm 11.1$ & $14.3 \pm 5.1$ \\
\rowcolor{mapleproposed}
\textbf{MAPLE-RF-S} & R & $4.85 \pm 0.58$ & \textcolor{maplesecond}{$0.82 \pm 0.28$} & \textcolor{maplebest}{$\mathbf{1.55 \pm 0.47}$} & \textcolor{maplesecond}{$53.4 \pm 12.0$} & \textcolor{maplebest}{$\mathbf{1.32 \pm 0.56}$} & $63.9 \pm 13.9$ & $14.3 \pm 5.1$ \\
\bottomrule
\end{tabular}
\par\vspace{1.2mm}\noindent
\begin{minipage}{0.99\textwidth}\footnotesize
C/R denote Clean/Robust training; H/S denote MAPLE-RF's Hard/Spatial variants. ES denotes energy score. LOCUS-DT uses a $49\times49$ grid unless marked $25^2$ or $13^2$; coarse scores are reconstructed on the original grid. Quality is mean $\pm$ sample standard deviation (SD) over 144 layout--map units, each averaging 48 RF snapshots. H100 times use three warmed runs per map/RX configuration; mean and SD are over 12 configurations. Timing includes visibility, transfers, fresh DT banks, coarse-score reconstruction, and posteriors; it excludes shared RF estimation and result storage. C/R models of each architecture share its R timing. Green marks MAPLE-RF-S (R); \textcolor{maplebest}{blue}/\textcolor{maplesecond}{orange} mark the best/second-best unrounded means. Definitions appear in Section~\ref{sec:quality_cost}.
\end{minipage}
\end{table*}

\section{Results and Discussion}
\label{sec:results}

\subsection{Single-Snapshot Localization across Map Coverage}
\label{sec:examples}
\label{sec:coverage_results}

We compare MAPLE-RF, partial-map LOCUS-DT, and Gaussian baselines across map coverage (Fig.~\ref{fig:localization_examples}), using identical map and RF observations within each row. Its last two columns isolate MAPLE-RF-S training regimes: under Moderate and Severe coverage, robust training places substantial mass near the TX, while clean training concentrates farther away. MAPLE-RF and LOCUS-DT follow observed geometry with irregular probability regions; Gaussian baselines produce smoother regions.

Fig.~\ref{fig:map_coverage_results} quantifies these trends with Mass NLL $-\log\widehat p_{\boldsymbol\omega,j_0}$ in nats, 1\,m recall (fraction of MAP estimates within 1\,m of the truth grid point $\bar{\mathbf x}_{j_0}^t$), and cone mass (probability within $\pm15^\circ$ of the bearing from the RX to $\mathbf x_0^t$). On paired Clean and Severe observations from the same layouts, MAPLE-RF-S (R) retains 94.3\% of its Clean-map recall, whereas MAPLE-RF-S (C) degrades as coverage decreases. Under robust training, Spatial supervision improves all three metrics over Hard. SG (R) and GMM4 (R) also vary little with coverage but trail MAPLE-RF-S (R) on all three metrics. MAPLE-RF-S (R) has the recall closest to LOCUS-DT (R) and slightly exceeds it in mean recall and cone mass on Severe maps, while LOCUS-DT (R) retains the lowest Mass NLL.

\begin{figure}[!t]
    \centering
    \includegraphics[trim={0 6bp 0 16bp},clip,width=0.96\columnwidth]{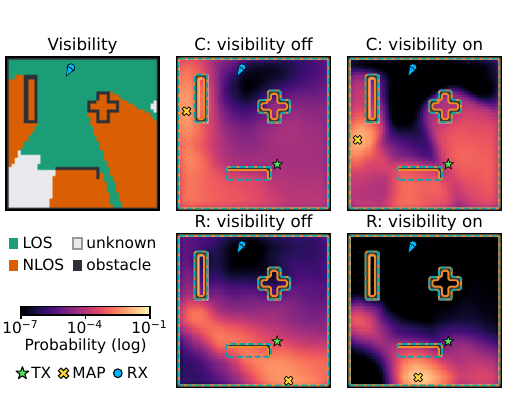}
    \par\vspace{1mm}
    \begingroup\footnotesize
\setlength{\tabcolsep}{1.5pt}
\renewcommand{\arraystretch}{0.98}
\begin{tabular}{p{\dimexpr.11\columnwidth-2\tabcolsep\relax}>{\centering\arraybackslash}p{\dimexpr.10\columnwidth-2\tabcolsep\relax}>{\centering\arraybackslash}p{\dimexpr.20\columnwidth-2\tabcolsep\relax}>{\centering\arraybackslash}p{\dimexpr.21\columnwidth-2\tabcolsep\relax}>{\centering\arraybackslash}p{\dimexpr.17\columnwidth-2\tabcolsep\relax}>{\centering\arraybackslash}p{\dimexpr.21\columnwidth-2\tabcolsep\relax}}
\toprule
 & & Clean & \multicolumn{3}{c}{Partially explored} \\
\cmidrule(lr){3-3}\cmidrule(lr){4-6}
Train & Vis. & \shortstack{1\,m mass\\{[\%]} $\uparrow$} & \shortstack{Mass NLL\\{[nat]} $\downarrow$} & \shortstack{ES\\{[m]} $\downarrow$} & \shortstack{1\,m mass\\{[\%]} $\uparrow$} \\
\midrule
C & Off & 25.7 & 7.32 & 1.95 & 15.8 \\
C & On & 44.9 & 8.38 & 2.47 & 11.6 \\
\addlinespace[1pt]
R & Off & 46.8 & 4.98 & 0.89 & 46.1 \\
\rowcolor[RGB]{237,246,231}
R & On & 55.5 & 4.85 & 0.82 & 53.4 \\
\bottomrule
\end{tabular}
\endgroup

    \caption{Visibility ablation of MAPLE-RF-S, using the same Mild example as Fig.~\ref{fig:localization_examples}. Solid orange and dashed cyan contours outline observed and true obstacles. Clean mass uses all Clean test maps; other summaries use the partial-map units of Table~\ref{tab:main_results_quality_cost}. Green marks MAPLE-RF-S (R).}
    \label{fig:visibility_ablation}
\end{figure}

\begin{figure}[!t]
    \centering
    \includegraphics[width=0.97\columnwidth]{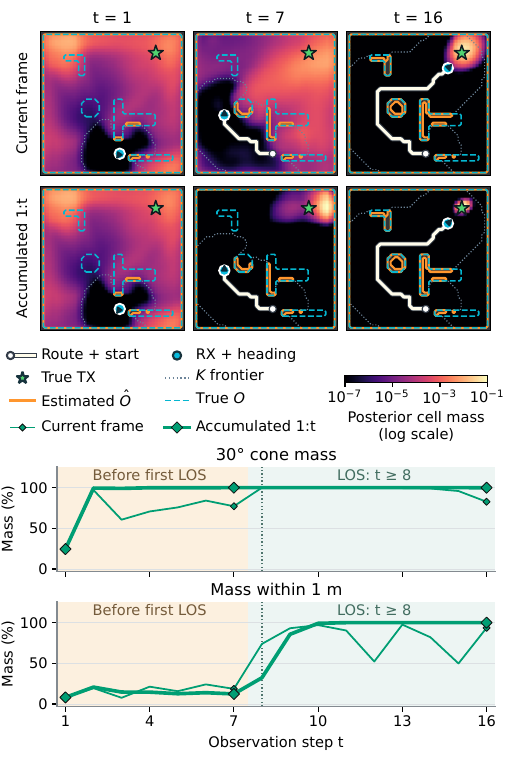}
    \caption{Current and accumulated MAPLE-RF-S (R) posteriors along one exploration trajectory. The current-snapshot MAP selects the route. Regional masses are compared at identical RX poses; the background transition marks the first true LOS observation. The 1\,m disk is centered on the continuous TX location.}
    \label{fig:snapshot_accumulation}
\end{figure}

\begin{figure}[!t]
    \centering
    \includegraphics[width=0.94\columnwidth]{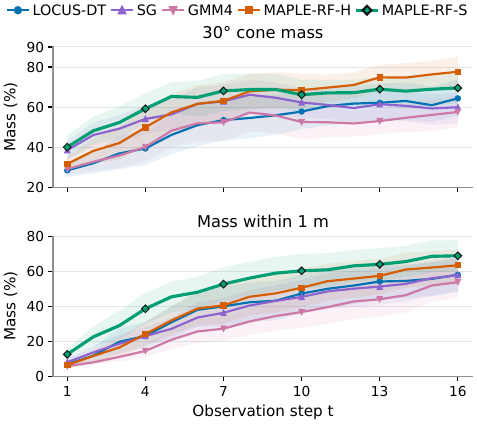}
    \caption{Accumulated regional probability mass for five base models with Robust (R) training on 100 shared, initially NLOS sequences across 34 held-out test layouts with four obstacles each. Each sequence contains 16 observations. Lines show equally weighted sequence means; shading gives pointwise 95\% confidence intervals from 2,000 shape-stratified layout-cluster bootstrap resamples. H/S denote Hard/Spatial supervision. The cone has a $30^\circ$ full angle; its axis starts at 20\%.}
    \label{fig:fusion_comparison}
\end{figure}

\subsection{Posterior Quality and Computational Cost}
\label{sec:quality_cost}

We assess posterior quality and query cost in Table~\ref{tab:main_results_quality_cost}, averaging the three partial-map levels equally. For masses $\widehat p_j$ on a method's native support $\mathcal C_{\mathrm{eval}}$, the expected distance is $D=\sum_{j\in\mathcal C_{\mathrm{eval}}}\widehat p_j\|\bar{\mathbf x}_j^t-\bar{\mathbf x}_{j_0}^t\|_2$, and the energy score (ES)~\cite{GneitingRaftery2007}
\begin{equation}
\mathrm{ES}=D-\frac12\sum_{j,k\in\mathcal C_{\mathrm{eval}}}\widehat p_j\widehat p_k\left\|\bar{\mathbf x}_j^t-\bar{\mathbf x}_k^t\right\|_2
\label{eq:energy_score}
\end{equation}
is a strictly proper score for the position distribution, measured in meters. The 1\,m mass sums $\widehat p_j$ over cells within 1\,m of $\bar{\mathbf x}_{j_0}^t$; unlike recall, it uses all probability near the TX. The MAP error is $\|\widehat{\mathbf x}^t-\bar{\mathbf x}_{j_0}^t\|_2$.

Robust training benefits MAPLE-RF-S much more than LOCUS-DT: partial-map 1\,m recall gains over Clean training are $50.7$ versus $11.8$ percentage points. Under robust training, MAPLE-RF-S has the lowest expected distance and nearly identical MAP error to full-grid LOCUS-DT. LOCUS-DT retains lower Mass NLL and ES, and higher 1\,m mass and recall. Within MAPLE-RF, Spatial supervision improves all six metrics over Hard under R training; MAPLE-RF-S (R) also outperforms SG (R) and GMM4 (R).

At batch size one, MAPLE-RF-S (R) computes the posterior about $218\times$ faster than full-grid LOCUS-DT (R), avoiding candidate-bank construction for each new map/RX configuration. Coarser DT grids reduce cost but progressively lose their quality advantage over MAPLE-RF-S (R). The $25\times25$ variant retains lower Mass NLL and higher recall than MAPLE-RF-S (R), while MAPLE-RF-S has lower ES, expected distance, and MAP error, and higher 1\,m mass. The $13\times13$ variant trails MAPLE-RF-S on all six metrics. MAPLE-RF-S remains about $63\times$ and $25\times$ faster than these two variants, respectively. All six quality rankings remain unchanged after restricting and renormalizing the seven robust variants' posteriors over their common candidate support.

The DT banks use general-purpose ray tracing. Because our simulations contain only direct and first-order specular paths, a specialized image-method solver could reduce bank construction time; the measured ratios do not represent the minimum achievable DT cost. MAPLE-RF avoids per-candidate simulation at inference, an advantage that may matter more with higher-order reflections, diffraction, or 3-D geometry.

\subsection{Visibility Ablation}
\label{sec:visibility_ablation}

We remove MAPLE-RF-S visibility by zeroing $B_{\mathrm{LOS}}$ and $B_{\mathrm{NLOS}}$ in otherwise identical training and testing. With visibility, Fig.~\ref{fig:visibility_ablation} shows sharper low-probability regions along confirmed LOS areas. Under C, visibility improves Clean-map quality but reduces partial-map quality; under R, it improves all six quality metrics on both Clean and partial maps.

\subsection{Posterior Fusion during Exploration}
\label{sec:accumulation}
\label{sec:fusion_comparison}

For MAPLE-RF, $T$ successive snapshots of a fixed scene and TX are fused by summing score vectors at common world-frame locations, $\widehat{\mathbf p}_{\boldsymbol\omega}^{1:T}=\operatorname{softmax}_{\mathcal C}(\sum_{s=1}^{T}\mathbf g_{\boldsymbol\omega,s})$, an empirical rule equal to the normalized product of snapshot posteriors. In Fig.~\ref{fig:snapshot_accumulation}, the RX updates the observed map with 2.5\,m scans and moves at most 1\,m through observed free space toward the current MAPLE-RF-S (R) MAP estimate, or toward a reachable frontier when that estimate is inaccessible. Current and accumulated predictions share the realized RX poses. Early high cone mass and low 1\,m mass indicate directional evidence with range uncertainty. Once LOS measurements arrive, the fused posterior concentrates at the true position and keeps nearly all its mass there, whereas the single-snapshot 1\,m mass continues to fluctuate.

Fig.~\ref{fig:fusion_comparison} compares five base models with R training on 100 initially NLOS sequences. All share observations, maps, and RX poses; routes follow MAPLE-RF-S (R) MAP estimates, which may favor that model. At step $T$, fused posteriors are normalized over the intersection of native supports across models and observations $1,\ldots,T$. Regional masses use the continuous TX location; cone mass is measured from the current RX pose. MAPLE-RF-S (R) has the highest mean accumulated 1\,m mass throughout and exceeds LOCUS-DT, SG, and GMM4 in both regional metrics at every step. MAPLE-RF-H (R) also finishes above these methods and has the highest cone mass late in the sequence; the cone constrains bearing, whereas the 1\,m disk also constrains range. Averaged over steps, MAPLE-RF-S places about $1.22\times$ as much probability within 1\,m as MAPLE-RF-H with comparable cone mass, favoring Spatial supervision for position localization.

\section{Conclusion}
\label{sec:conclusion}

We compared two posterior estimators for single-snapshot RF localization on partial maps: an extension of LOCUS-DT and MAPLE-RF, which scores all candidates in one U-Net pass without simulating propagation at inference. Mixed-coverage training raised DT partial-map 1\,m recall from 54.7\% to 66.4\%; MAPLE-RF retained about 94\% of its complete-map recall with 75--85\% of the map unobserved. On partial maps, full-grid DT led on four quality metrics; MAPLE-RF achieved nearly identical MAP error and lower expected distance with about $218\times$ lower fresh-query cost than full-grid DT using general-purpose ray tracing. It outperformed an inference-only $13\times13$ DT grid on all six metrics at about $25\times$ lower cost. Accumulated MAPLE-RF posteriors placed the most probability within 1\,m of the source along exploration trajectories.

The evaluation assumes first-order paths and noise-free observed maps and RX poses; a specialized DT solver could narrow the timing gap. Fusion ignores observation correlations and uses routes selected by MAPLE-RF. Future work will validate both approaches with real measurements and SLAM-estimated maps and poses, extend them to multi-room environments with varied sizes and obstacle shapes, higher-order reflections, and diffraction, account for correlated observations in fusion, and use posteriors in active source-seeking planners.

\end{NoHyper}

\begin{thebibliography}{10}
\providecommand{\url}[1]{#1}
\csname url@rmstyle\endcsname
\providecommand{\newblock}{\relax}
\providecommand{\bibinfo}[2]{#2}
\providecommand\BIBentrySTDinterwordspacing{\spaceskip=0pt\relax}
\providecommand\BIBentryALTinterwordstretchfactor{4}
\providecommand\BIBentryALTinterwordspacing{\spaceskip=\fontdimen2\font plus
\BIBentryALTinterwordstretchfactor\fontdimen3\font minus
  \fontdimen4\font\relax}
\providecommand\BIBforeignlanguage[2]{{%
\expandafter\ifx\csname l@#1\endcsname\relax
\typeout{** WARNING: IEEEtran.bst: No hyphenation pattern has been}%
\typeout{** loaded for the language `#1'. Using the pattern for}%
\typeout{** the default language instead.}%
\else
\language=\csname l@#1\endcsname
\fi
#2}}

\bibitem{SpotFi2015}
M.~Kotaru, K.~Joshi, D.~Bharadia, and S.~Katti, ``{SpotFi}: Decimeter level
  localization using {WiFi},'' in \emph{Proc. ACM SIGCOMM}, 2015, pp. 269--282.

\bibitem{DLoc2020}
R.~Ayyalasomayajula, A.~Arun, C.~Wu, S.~Sharma, A.~R. Sethi, D.~Vasisht, and
  D.~Bharadia, ``Deep learning based wireless localization for indoor
  navigation,'' in \emph{Proc. ACM Int. Conf. Mobile Computing and Networking
  (MobiCom)}, 2020, pp. 214--227.

\bibitem{Yin2022}
M.~Yin, A.~K. Veldanda, A.~Trivedi, J.~Zhang, K.~Pfeiffer, Y.~Hu, S.~Garg,
  E.~Erkip, L.~Righetti, and S.~Rangan, ``Millimeter wave wireless assisted
  robot navigation with link state classification,'' \emph{IEEE Open J. Commun.
  Soc.}, vol.~3, pp. 493--507, 2022.

\bibitem{Yin2024}
M.~Yin, T.~Li, H.~Lei, Y.~Hu, S.~Rangan, and Q.~Zhu, ``Zero-shot wireless
  indoor navigation through physics-informed reinforcement learning,'' in
  \emph{Proc. IEEE Int. Conf. Robot. Autom. (ICRA)}, 2024, pp. 5111--5118.

\bibitem{Twigg2012}
J.~N. Twigg, J.~R. Fink, P.~L. Yu, and B.~M. Sadler, ``{RSS} gradient-assisted
  frontier exploration and radio source localization,'' in \emph{Proc. IEEE
  Int. Conf. Robot. Autom. (ICRA)}, 2012, pp. 889--895.

\bibitem{Denniston2023}
C.~E. Denniston, O.~Peltzer, J.~Ott, S.~Moon, S.-K. Kim, G.~S. Sukhatme, M.~J.
  Kochenderfer, M.~Schwager, and A.-a. Agha-mohammadi, ``Fast and scalable
  signal inference for active robotic source seeking,'' in \emph{Proc. IEEE
  Int. Conf. Robot. Autom. (ICRA)}, 2023, pp. 7909--7915.

\bibitem{RadioUNet2021}
R.~Levie, {\c{C}}.~Yapar, G.~Kutyniok, and G.~Caire, ``{RadioUNet}: Fast radio
  map estimation with convolutional neural networks,'' \emph{IEEE Trans.
  Wireless Commun.}, vol.~20, no.~6, pp. 4001--4015, 2021.

\bibitem{LocUNet2023}
{\c{C}}.~Yapar, R.~Levie, G.~Kutyniok, and G.~Caire, ``Real-time outdoor
  localization using radio maps: A deep learning approach,'' \emph{IEEE Trans.
  Wireless Commun.}, vol.~22, no.~12, pp. 9703--9717, 2023.

\bibitem{SionnaRT2025}
F.~{A{\"i}t Aoudia}, J.~Hoydis, M.~Nimier-David, B.~Nicolet, S.~Cammerer, and
  A.~Keller, ``{Sionna RT}: Technical report,'' 2025, arXiv:2504.21719.

\bibitem{Li2025DTWIN}
T.~Li, H.~Lei, H.~Guo, M.~Yin, Y.~Hu, Q.~Zhu, and S.~Rangan, ``Digital
  twin-enhanced wireless indoor navigation: Achieving efficient environment
  sensing with zero-shot reinforcement learning,'' \emph{IEEE Open J. Commun.
  Soc.}, vol.~6, pp. 2356--2372, 2025.

\bibitem{LOCUSDT}
H.~Lei, R.~Bomfin, M.~Chafii, and S.~Rangan, ``{LOCUS-DT}: Localization via
  observation-conditioned uncertainty scoring with digital twins,'' 2026,
  arXiv:2608.00406.

\bibitem{Yamauchi1997}
B.~Yamauchi, ``A frontier-based approach for autonomous exploration,'' in
  \emph{Proc. IEEE Int. Symp. Comput. Intell. Robot. Autom. (CIRA)}, 1997, pp.
  146--151.

\bibitem{Cadena2016}
C.~Cadena, L.~Carlone, H.~Carrillo, Y.~Latif, D.~Scaramuzza, J.~Neira, I.~Reid,
  and J.~J. Leonard, ``Past, present, and future of simultaneous localization
  and mapping: Toward the robust-perception age,'' \emph{IEEE Trans. Robot.},
  vol.~32, no.~6, pp. 1309--1332, 2016.

\bibitem{Gonultas2022}
E.~G{\"o}n{\"u}lta{\c{s}}, E.~Lei, J.~Langerman, H.~Huang, and C.~Studer,
  ``{CSI}-based multi-antenna and multi-point indoor positioning using
  probability fusion,'' \emph{IEEE Trans. Wireless Commun.}, vol.~21, no.~4,
  pp. 2162--2176, 2022.

\bibitem{Leitinger2019}
E.~Leitinger, F.~Meyer, F.~Hlawatsch, K.~Witrisal, F.~Tufvesson, and M.~Z. Win,
  ``A belief propagation algorithm for multipath-based {SLAM},'' \emph{IEEE
  Trans. Wireless Commun.}, vol.~18, no.~12, pp. 5613--5629, 2019.

\bibitem{MCCLE2025}
H.~Lei, H.~Guo, T.~Svensson, and S.~Rangan, ``Beyond point estimates:
  Likelihood-based full-posterior wireless localization,'' 2025,
  arXiv:2509.25719.

\bibitem{Charrow2014}
B.~Charrow, N.~Michael, and V.~Kumar, ``Cooperative multi-robot estimation and
  control for radio source localization,'' \emph{Int. J. Robot. Res.}, vol.~33,
  no.~4, pp. 569--580, 2014.

\bibitem{Kim2025}
S.~Kim, S.~Moon, I.~Petrunin, H.-S. Shin, and S.~Khattak, ``Autonomous robotic
  radio source localization via a novel {G}aussian mixture filtering
  approach,'' in \emph{Proc. Int. Conf. Inf. Fusion (FUSION)}, 2025, pp. 1--8.

\bibitem{Li2025PiPRL}
\BIBentryALTinterwordspacing
T.~Li, H.~Lei, M.~Yin, and Y.~Hu, ``Reinforcement learning with
  physics-informed symbolic program priors for zero-shot wireless indoor
  navigation,'' in \emph{Inductive Biases in Reinforcement Learning (IBRL)
  Workshop at RLC}, 2025. [Online]. Available:
  \url{https://openreview.net/forum?id=w1Lg5cxCCU}
\BIBentrySTDinterwordspacing

\bibitem{Clark2022}
L.~Clark, J.~Edlund, M.~S. Net, T.~S. Vaquero, and A.-a. Agha-mohammadi,
  ``{PropEM-L}: Radio propagation environment modeling and learning for
  communication-aware multi-robot exploration,'' in \emph{Proc. Robot.: Sci.
  Syst. (RSS)}, 2022.

\bibitem{Kanhere2025}
O.~Kanhere and T.~S. Rappaport, ``Map-assisted millimeter wave and terahertz
  position location and sensing,'' \emph{IEEE Trans. Wireless Commun.},
  vol.~24, no.~6, pp. 5323--5336, 2025.

\bibitem{Leitinger2015}
E.~Leitinger, P.~Meissner, M.~Lafer, and K.~Witrisal, ``Simultaneous
  localization and mapping using multipath channel information,'' in
  \emph{Proc. IEEE Int. Conf. Commun. Workshop (ICCW)}, 2015, pp. 754--760.

\bibitem{Ulmschneider2023}
M.~Ulmschneider and C.~Gentner, ``User tracking with multipath assisted
  positioning-based fingerprinting and deep learning,'' in \emph{Proc. Eur.
  Conf. Antennas Propag. (EuCAP)}, 2023, pp. 1--5.

\bibitem{Ulmschneider2024}
M.~Ulmschneider, C.~Gentner, and A.~Dammann, ``Mixture density networks for
  multipath assisted positioning-based fingerprinting,'' in \emph{Proc. Eur.
  Conf. Antennas Propag. (EuCAP)}, 2024, pp. 1--5.

\bibitem{UNet2015}
O.~Ronneberger, P.~Fischer, and T.~Brox, ``{U-Net}: Convolutional networks for
  biomedical image segmentation,'' in \emph{Proc. Int. Conf. Medical Image
  Computing and Computer-Assisted Intervention (MICCAI)}, 2015, pp. 234--241.

\bibitem{Attention2017}
A.~Vaswani, N.~Shazeer, N.~Parmar, J.~Uszkoreit, L.~Jones, A.~N. Gomez,
  {\L}.~Kaiser, and I.~Polosukhin, ``Attention is all you need,'' in
  \emph{Advances in Neural Information Processing Systems}, vol.~30, 2017, pp.
  5998--6008.

\bibitem{Bomfin2025}
R.~Bomfin, A.~Rasteh, A.~Bazzi, H.~Guo, H.~Lee, M.~Mezzavilla, S.~Rangan,
  J.~Choi, and M.~Chafii, ``Multi-band channel sensing in the upper mid-band
  ({FR3}),'' in \emph{Proc. IEEE Global Commun. Conf. (GLOBECOM)}, 2025, pp.
  5127--5132.

\bibitem{GneitingRaftery2007}
T.~Gneiting and A.~E. Raftery, ``Strictly proper scoring rules, prediction, and
  estimation,'' \emph{J. Amer. Statist. Assoc.}, vol. 102, no. 477, pp.
  359--378, 2007.

\end{thebibliography}
\end{document}